\documentclass[runningheads]{llncs}
\usepackage[T1]{fontenc}

\usepackage{amsmath}
\usepackage{amssymb}
\usepackage{graphicx}
\usepackage{hyperref}
\usepackage{array,tabularx,graphicx,makecell}
\newcolumntype{M}[1]{>{\centering\arraybackslash}m{#1}}

\begin{document}
\title{Iterative Layer-wise Distillation for Efficient Compression of Large Language Models}
\titlerunning{Iterative Layer-wise Distillation}
%
\author{Grigory Kovalev\inst{1}\orcidID{0009-0008-5356-4527} \and
Mikhail Tikhomirov\inst{1}\orcidID{0000-0001-7209-9335}}
\authorrunning{G. Kovalev, M. Tikhomirov}
%
\institute{Lomonosov Moscow State University, Russia \\
\email{kaengreg@ya.ru \hspace{1mm} tikhomirov.mm@gmail.com}}

\maketitle              

\begin{abstract}
This work investigates distillation methods for large language models (LLMs) with the goal of developing compact models that preserve high performance. Several existing approaches are reviewed, with a discussion of their respective strengths and limitations. An improved method based on the ShortGPT approach has been developed, building upon the idea of incorporating iterative evaluation of layer importance. At each step, importance is assessed by measuring performance degradation when individual layers are removed, using a set of representative datasets. This process is combined with further training using a joint loss function based on KL divergence and mean squared error. Experiments on the Qwen2.5-3B model show that the number of layers can be reduced from 36 to 28 (resulting in a 2.47 billion parameter model) with only a 9.7\% quality loss, and to 24 layers with an 18\% loss. The findings suggest that the middle transformer layers contribute less to inference, underscoring the potential of the proposed method for creating efficient models. The results demonstrate the effectiveness of iterative distillation and fine-tuning, making the approach suitable for deployment in resource-limited settings.

\end{abstract}

\section{Introduction}
\qquad Recent advances in natural language processing (NLP) have been driven by large language models (LLMs) such as GPT \cite{brown2020language} and BERT \cite{devlin-etal-2019-bert}, built on the transformer architecture \cite{vaswani2017attention} and trained on massive corpora like Common Crawl\footnote{\url{https://commoncrawl.org/}} and Wikipedia\footnote{\url{https://www.wikipedia.org/}}. These models have achieved remarkable performance in tasks ranging from text generation \cite{chiang2024chatbot}, machine translation \cite{costa2022no}, and question answering \cite{rajpurkar2016squad,clark2020tydi}, to information retrieval \cite{bajaj2016ms,kwiatkowski2019natural,zhang2023miracl}.

Despite their success, LLMs are computationally expensive, requiring significant memory and computational resources, which poses challenges for deployment on mobile, embedded, or other resource-constrained devices. Their size - often tens of billions of parameters - limits real-world use where efficiency is crucial.

\begin{figure}[h]
\centering
\includegraphics[width=1\linewidth]{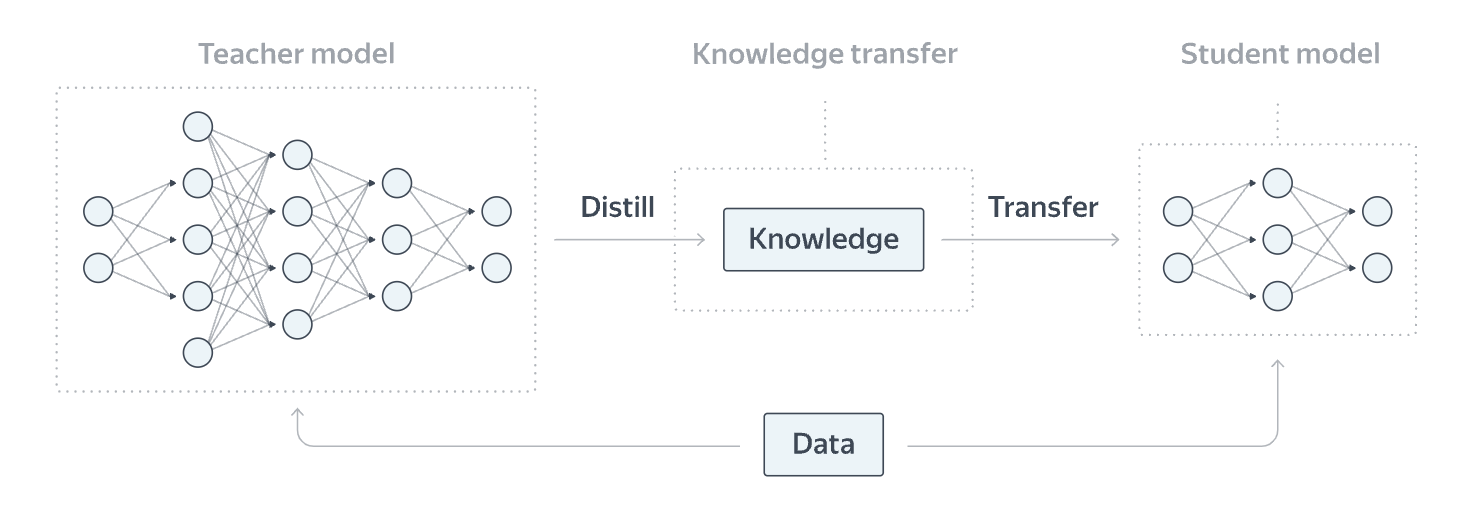}
\caption{General scheme of model distillation \cite{gou2021knowledge}}
\label{fig:model-distil}
\end{figure}

Model distillation addresses this challenge by transferring the knowledge of a large teacher model to a smaller student model, as illustrated in Figure \ref{fig:model-distil} \cite{yandexHandbook}. 

In this process, the student learns not only from original data but also from the soft labels (probabilistic outputs) of the teacher, which encode richer information than hard labels. This enables the student to generalize better and be more robust, while reducing the model’s size and inference cost.

Distilled models thus combine strong performance with efficiency, making LLM's accessible on devices with limited resources, such as smartphones and IoT devices. 

In this work, we address the challenge of compressing large language models by developing an improved distillation framework that combines iterative, data-driven evaluation of layer importance with targeted quality restoration techniques. Our method systematically identifies and removes less important transformer layers based on data-driven performance assessment and employs a combined loss function to fine-tune the reduced model on both the logits and hidden states of the original model. Experiments on the Qwen2.5-3B\footnote{https://huggingface.co/Qwen/Qwen2.5-3B} model demonstrate that this approach enables substantial reductions in model size - removing up to a third of layers with only a minor drop in quality - thus making high-performing language models more suitable for deployment in resource-constrained environments. These results underscore the effectiveness of iterative distillation and pave the way for further advances in efficient large language model adaptation. To facilitate reproducibility and future research, we publicly release the full implementation at GitHub\footnote{https://github.com/kaengreg/layer-wise\_distillation}.

\section{Related Work}

\subsection{Knowledge Distillation}
Knowledge distillation in its modern form was introduced by Hinton et al.\cite{hinton2015distilling}, proposing that a compact student model can be trained to mimic a large teacher model by using the teacher’s predicted probabilities (soft labels). The distillation objective combines standard cross-entropy with a Kullback-Leibler divergence term, encouraging the student not only to match hard labels but also the probability distribution produced by the teacher:
\[
\mathcal{L}_{KD} = \frac{1}{N} \sum_{i=1}^N \left( - \sum_{j=1}^K y_{ij} \log p_{ij} + \lambda D_{KL}(\mathbf{p}_i \| \mathbf{q}_i) \right)
\]
where $y_{ij}$ is the hard label, $p_{ij}$ and $q_{ij}$ are the student and teacher class probabilities, respectively. This “dark knowledge” in the teacher’s output captures class relationships, providing richer supervision than hard labels alone.

\subsection{Distilling Step-by-Step}
The ``Distilling Step-by-Step'' approach \cite{hsieh2023distilling} extends classical distillation by leveraging large language models (LLMs) as teachers to provide not only labels, but also natural language  rationales - explanations justifying the predictions. Inspired by Chain-of-Thought prompting \cite{wei2022chain}, the teacher generates both pseudo-labels and explanations for each input. The student is then trained on triplets $(x^p, r^p, y^p)$: input, rationale, and label, which enhances learning, especially in low-resource settings.

\begin{figure}[h!]
    \centering
    \includegraphics[width=1\linewidth]{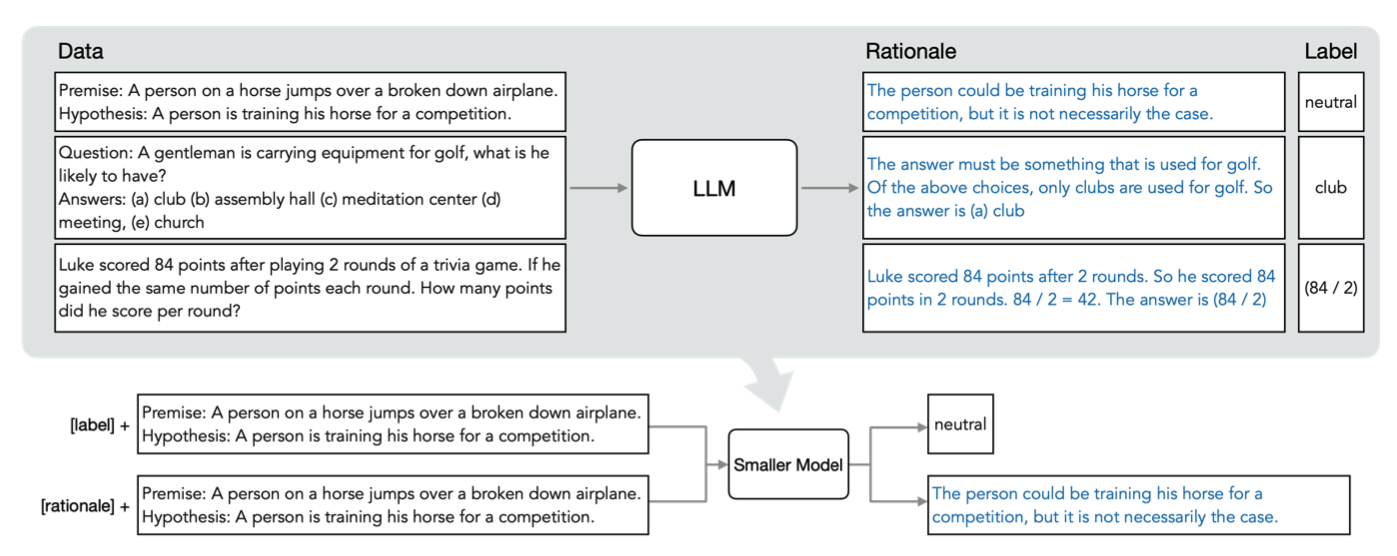}
    \caption{An example of how teacher labels and explanations are used during student training \cite{hsieh2023distilling}.}
    \label{fig:distil-sbs-exp}
\end{figure}

The method involves two steps: generating teacher rationales for unlabeled data, and training the student using both text and rationale as input. The loss combines label and rationale generation:
\[
\mathcal{L} = \mathcal{L}_{\text{label}} + \lambda \mathcal{L}_{\text{rationale}},
\]
where $\lambda$ controls the importance of the rationale term. At inference, the student predicts without access to rationales, improving efficiency.

\begin{figure}[h]
    \centering
    \includegraphics[width=0.75\linewidth]{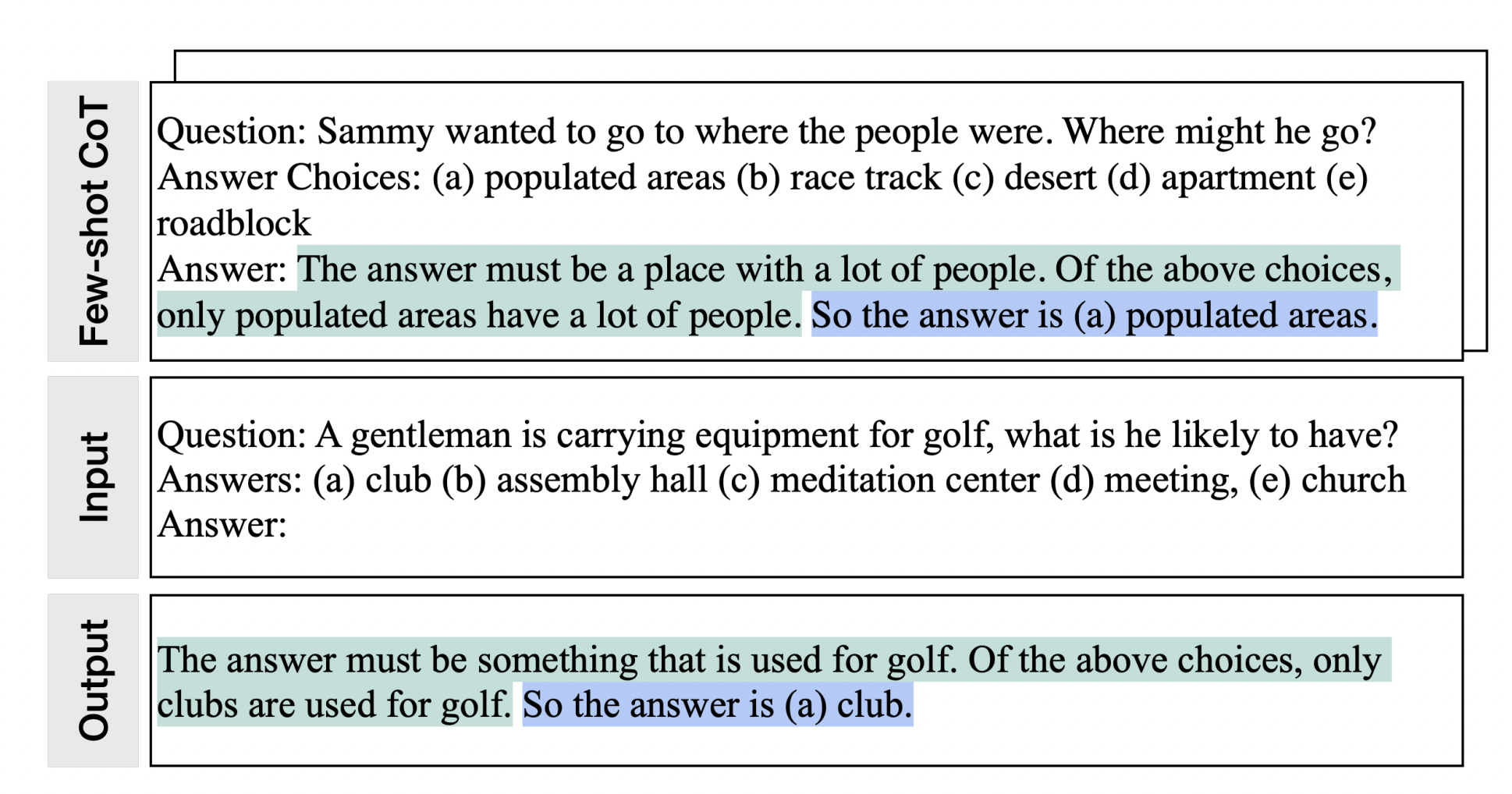}
    \caption{Illustration for constructing input-rationale-label triplets \cite{hsieh2023distilling}.}
    \label{fig:triplets}
\end{figure}

\subsection{MiniLLM}
MiniLLM \cite{gu2023minillm} refines knowledge distillation for generation tasks by focusing on the reverse Kullback-Leibler divergence $KL[q_\theta\|p]$ instead of the forward $KL[p\|q]$. While the standard approach forces the student to cover all teacher modes (sometimes yielding unlikely generations), minimizing the reverse KL encourages the student to focus on the teacher’s main modes and avoid improbable outputs. The reverse KL is minimized with a policy gradient method \cite{sutton1999policy}, and training is stabilized using one-step decomposition, mixed teacher forcing, and length normalization.

\begin{figure*}[h!]
    \centering
    \includegraphics[width=0.9\linewidth]{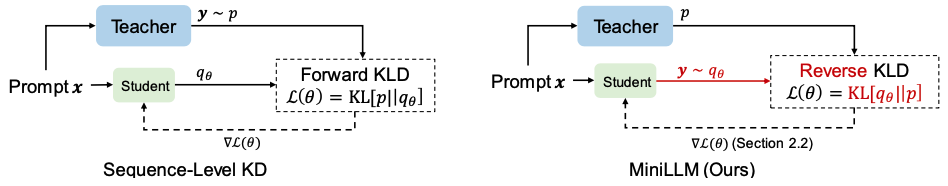}
    \caption{Scheme of distillation using reverse-KLD \cite{gu2023minillm}.}
    \label{fig:rev-kld}
\end{figure*}

This approach produces student models that are more efficient and accurate for text generation.

\subsection{Distillation via Linear Pruning}

The authors \cite{razzhigaev2024your} propose an innovative distillation approach that exploits the linearity inherent in transformer architectures. By identifying and pruning highly linear layers in transformer architectures, substituting them with linear approximations, and employing mean squared error (MSE) layerwise distillation, this method ensures the distilled model retains strong performance despite significant pruning. Evaluations on datasets such as WikiText\cite{merity2016pointer} and ARC-easy\cite{allenai:arc} demonstrate that this technique effectively reduces performance degradation associated with layer removal, providing a robust strategy for developing efficient transformer models.

\subsection{ShortGPT}
ShortGPT \cite{men2024shortgpt} focuses on reducing model redundancy by identifying less influential transformer layers using the Block Influence (BI) metric:
\[
\mathrm{BI}_i = 1 - \mathbb{E}_{X,t}\frac{X_{i,t}^T X_{i+1,t}}{\lVert X_{i,t}\rVert_2 \lVert X_{i+1,t}\rVert_2}
\]
where $X_{i,t}$ is the hidden state vector of token $t$ at the $i^{th}$ layer. A low BI score indicates that $X_{i,t}$ and $X_{i+1,t}$ have high cosine similarity, meaning the layer produces only minimal transformations. Such layer is therefore considered redundant and more suitable for pruning.
The method demonstrates that removing layers with low BI can significantly shrink models like Llama2-7B \cite{touvron2023llama} and Baichuan2-7B \cite{yang2023baichuan} with minimal quality loss.

\begin{figure}[!h]
    \centering
    \includegraphics[width=0.5\linewidth]{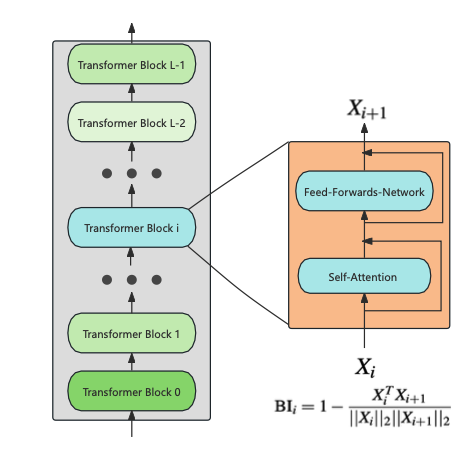}
    \caption{Block Influence (BI) metric for transformer layers \cite{men2024shortgpt}.}
    \label{fig:bi}
\end{figure}

This metric guides pruning, reducing computational costs and enabling efficient deployment of LLMs.


In summary, these recent advances in distillation - leveraging soft-labels, rationales, modified loss functions, and structural metrics - enable the creation of compact, high-quality models suitable for efficient deployment.

\section{Methodology: Iterative Evaluation and Distillation}
\subsection{Improvements to the ShortGPT Approach}
    $\qquad$The ShortGPT method, described in \cite{men2024shortgpt}, involves evaluating the importance of layers only for the original model. This approach can be improved by measuring layer importance iteratively after each layer removal, since such an operation can affect the model’s internal structure and, consequently, the generation process in which the Block Influence metric is calculated.


    Fine-tuning can be performed in the following ways:
    \begin{enumerate}
        \item Fine-tuning the student model on the logits of the teacher model.
        \item Fine-tuning the student model using the hidden states of the teacher model.
        \item Fine-tuning the student model simultaneously on the logits and hidden states of the teacher model.
        \item Fine-tuning for a specialized task or domain.
    \end{enumerate}
    
    In these and subsequent formulations, the teacher model refers to the original model, and the student model is the model, initially the same model as a teacher, from which $k$ layers have been removed.

    \noindent
    In this work, two modifications of the ShortGPT approach are investigated and compared:
    \begin{enumerate}
        \item Iterative measurement of layer importance after each layer removal.
        \item Iterative measurement of layer importance after each layer removal, with additional fine-tuning of the model on the outputs of the teacher model to restore quality.
    \end{enumerate}
        
\subsection{A New Approach to Layer Importance Evaluation}
    $\qquad$The evaluation of layer importance in ShortGPT \cite{men2024shortgpt} is based on the Block Influence metric, but it does not always correlate well with the actual model quality.
    Therefore, the following algorithm is proposed to provide a more objective assessment of layer importance:
    \begin{enumerate}
      \item Using the current model with $n$ layers, form $n$ variants by removing one layer at a time.
      \item Each model is tested on the selected representative datasets, applying metrics such as F-Score \cite{sokolova2006beyond}, ROUGE \cite{lin2004rouge}; for each layer, the drop in quality is recorded.
      \item Layers whose removal results in minimal quality loss are considered the least important.
    \end{enumerate}

    \noindent This approach, as well as the modified ShortGPT method, is analyzed in two variants:
    \begin{enumerate}
        \item Iterative determination of layer importance after their sequential removal.
        \item Iterative determination of layer importance with subsequent fine-tuning on the outputs of the teacher model to improve student model quality.
    \end{enumerate}

\section{Datasets}
\subsection{Evaluation Datasets}
    
    In this work, seven representative datasets were selected to assess layer importance. These datasets cover a range of natural language processing domains, including comprehension, text generation, translation, and general knowledge evaluation. This selection allows for a thorough and objective evaluation of model quality following layer removal.  The datasets are introduced below, with each chosen to highlight a different capability of the model:
        
   \begin{enumerate}
    \item \textbf{enMMLU (English Massive Multitask Language Understanding) \cite{hendrycks2020measuring}}: This dataset consists of multiple-choice questions in English, covering various areas of knowledge, including natural sciences, humanities, and professional skills. It is designed to test a model’s ability to understand and utilize knowledge in different contexts, making it useful for assessing the model’s general knowledge.

    \item \textbf{ruMMLU (Russian Massive Multitask Language Understanding)}: This dataset is similar to enMMLU but adapted for the Russian language and includes multiple-choice questions. It is important for testing models in Russian-language environments.

    \item \textbf{treewayabstractive}: This dataset is designed for abstractive summarization tasks - the model must process a text in Russian and compose its summary. This allows the assessment of how well the model generates coherent and informative text.

    \item \textbf{cp\_doc\_ru}: This dataset is used for document copying tasks in Russian. The model must reproduce the entire document, which helps assess its accuracy when copying text.

    \item \textbf{cp\_para\_ru}: Similar to \texttt{cp\_doc\_ru}, this dataset is for paragraph-level copying tasks in Russian. It assesses the model’s ability to reproduce text at the paragraph level.

    \item \textbf{flores\_en\_ru \cite{costa2022no}}: This dataset includes sentence pairs for translation from English to Russian. It is intended to test machine translation quality.

    \item \textbf{flores\_ru\_en}: The counterpart of \texttt{flores\_en\_ru}, but for translation from Russian to English. This dataset is important for evaluating bidirectional translation quality and the model’s ability to handle Russian as the source language.
\end{enumerate}

The model’s overall performance is reported as the Aggregate Score, defined as the average of its results across all selected datasets. This provides a single number for comparison, but the detailed per-dataset results for the main considered method are also presented in the Table \ref{tab:full-results}.

\subsection{Dataset for Distillation}
  Selecting a diverse and high-quality dataset for the fine-tuning stage is a critical factor in achieving superior model performance. For this study, we adopted the dataset utilized in \cite{tikhomirov2024facilitating}, which is constructed upon the comprehensive \texttt{IlyaGusev/rulm} \footnote{https://huggingface.co/datasets/IlyaGusev/rulm} collection. This dataset draws from a broad spectrum of sources, including Russian and English Wikipedia articles, literary works, social media content, and reputable news outlets. Such diversity and quality ensure a robust and representative data foundation, facilitating effective knowledge transfer and enhancing the outcomes of the distillation process.

\section{Experiments}

  \subsection{Experiments with Loss Functions and Hyperparameter Tuning}

To determine the most effective loss functions for the distillation stage following layer pruning, we conducted a series of experiments with various configurations. We evaluated \texttt{KLDivLoss}\footnote{https://docs.pytorch.org/docs/stable/generated/torch.nn.KLDivLoss.html} for aligning logits and probability distributions, as well as \texttt{MSELoss}\footnote{https://pytorch.org/docs/stable/generated/torch.nn.MSELoss.html} for matching hidden states and intermediate outputs. Both individual and combined loss functions were systematically assessed to identify which best preserved model quality during distillation.

\subsubsection{Logit-based Fine-tuning with KLDivLoss}
$\qquad$In logit-based fine-tuning, the student model is trained to reproduce the probability distribution produced by the teacher. Logits are the raw outputs of the final model layer prior to activation, encoding the predicted probabilities for each class. The Kullback-Leibler divergence (\texttt{KLDivLoss}) minimizes the divergence between the student's and teacher's output distributions.

\begin{figure}
    \centering
    \includegraphics[width=1\linewidth]{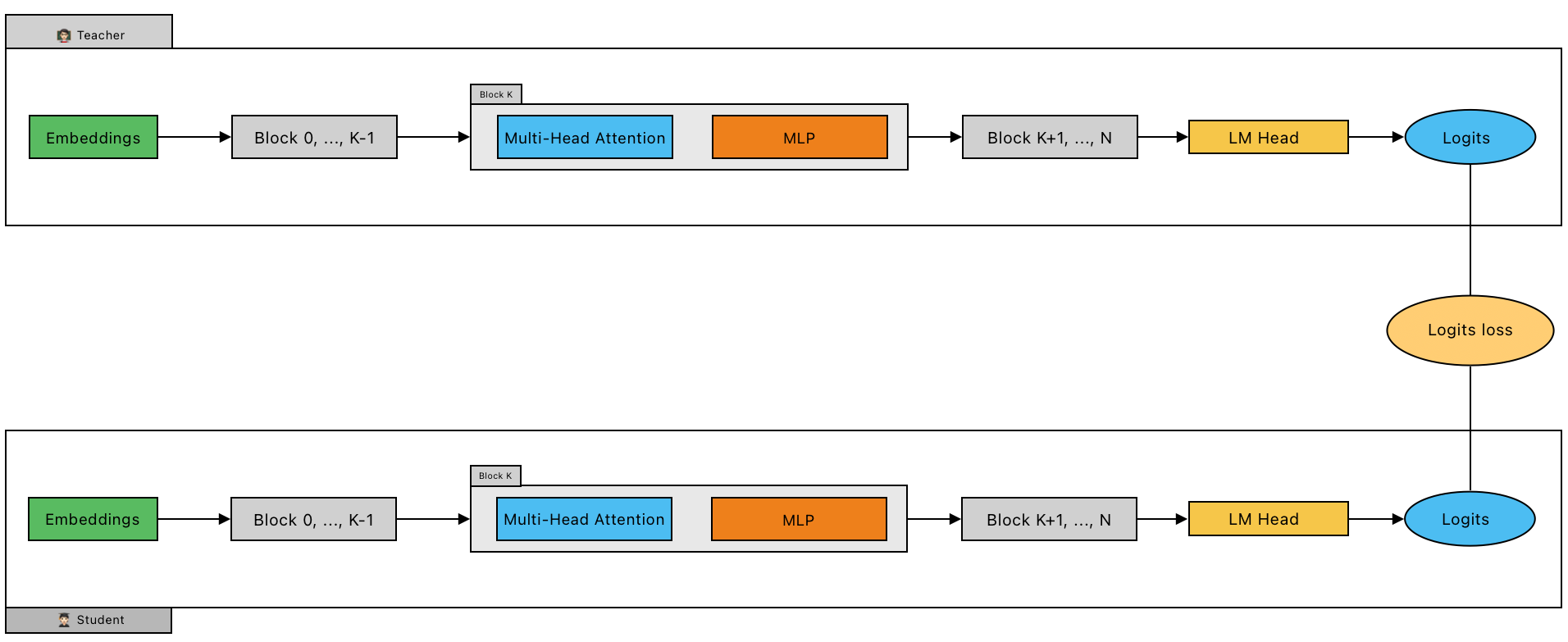}
    \caption{Logit-based fine-tuning}
    \label{fig:logits-ft}
\end{figure}

This approach enables the student not only to replicate the teacher’s predictions but also to capture the structural and semantic information present in the teacher’s distribution. The scheme is illustrated in Figure \ref{fig:logits-ft}.

In several experiments, log-probabilities (logprobs) - obtained from applying softmax to logits - were used instead of raw logits. Logprobs, being the logarithms of class probabilities, are particularly suitable for \texttt{KLDivLoss}, which expects probability distributions as input. Using logprobs offers:
\begin{enumerate}
    \item Simplified computations, as softmax is already applied
    \item Reduced influence of extreme logit values, enhancing training stability
\end{enumerate}

Additionally, we experimented with both forward and reverse KL divergence - following the recommendation of MiniLLM \cite{gu2023minillm} - and used log-probabilities as targets to examine how different probability representations influence fine-tuning quality.

\subsubsection{Fine-tuning on Hidden States with MSELoss}
\begin{figure}
    \centering
    \includegraphics[width=1\linewidth]{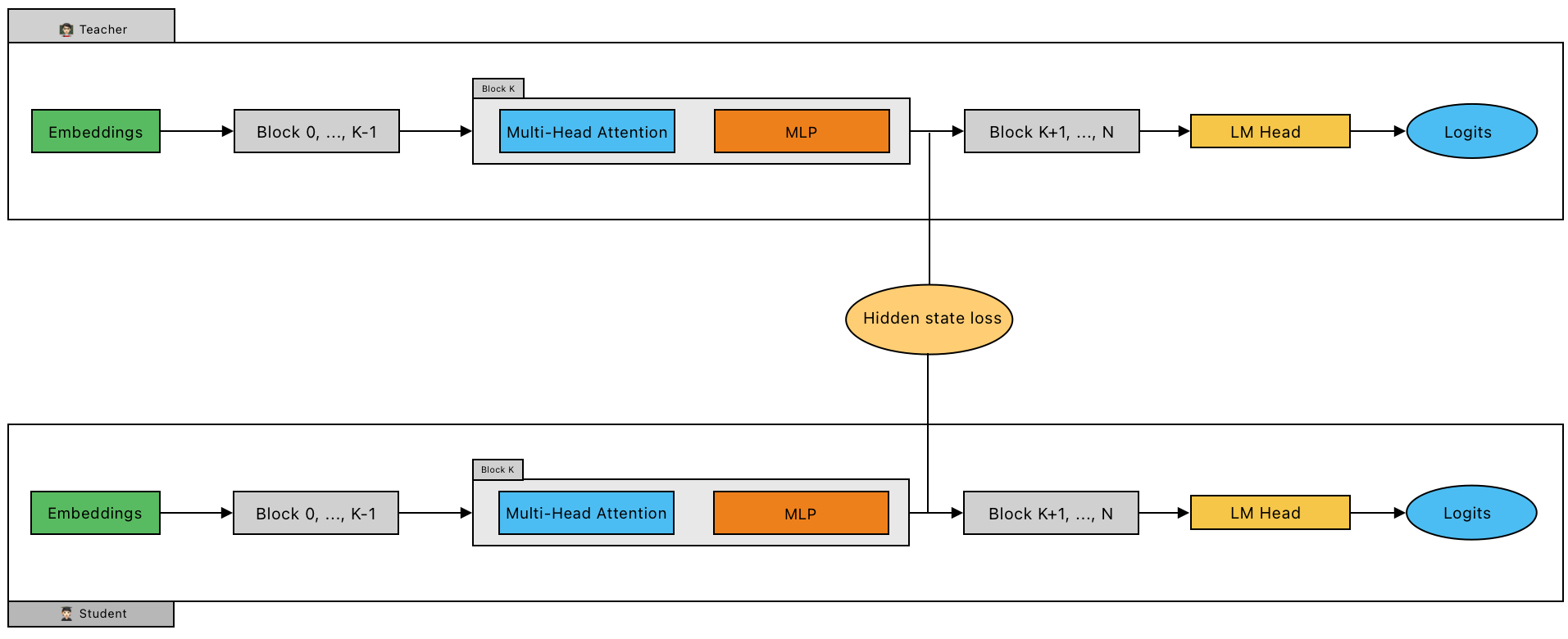}
    \caption{Fine-tuning on hidden states}
    \label{fig:hs-ft}
\end{figure}

$\qquad$For knowledge transfer via intermediate representations, we used \texttt{MSELoss} (mean squared error) as a distance measure to align hidden states between teacher and student (see Fig. \ref{fig:hs-ft}). MSE penalizes large deviations between activation vectors, providing a simple and stable objective for training. Hidden states, the activations at each layer, contain valuable information about the model's internal structures and semantic relationships. Aligning them encourages the student to reproduce these representations, facilitating deeper knowledge transfer.

We explored the following MSELoss variants:
\begin{enumerate}
    \item Between outputs of trainable layers: minimizes the discrepancy between corresponding activations in student and teacher, helping the student retain local processing features.
    \item Between outputs of the last layers: aligns the final representations before logits formation.
    \item Between outputs of the last trainable layers.
    \item Combined: sums MSELoss across both last trainable and last layers, unifying local and global representation optimization.
\end{enumerate}

\begin{figure}[h!]
    \centering
    \includegraphics[width=1\linewidth]{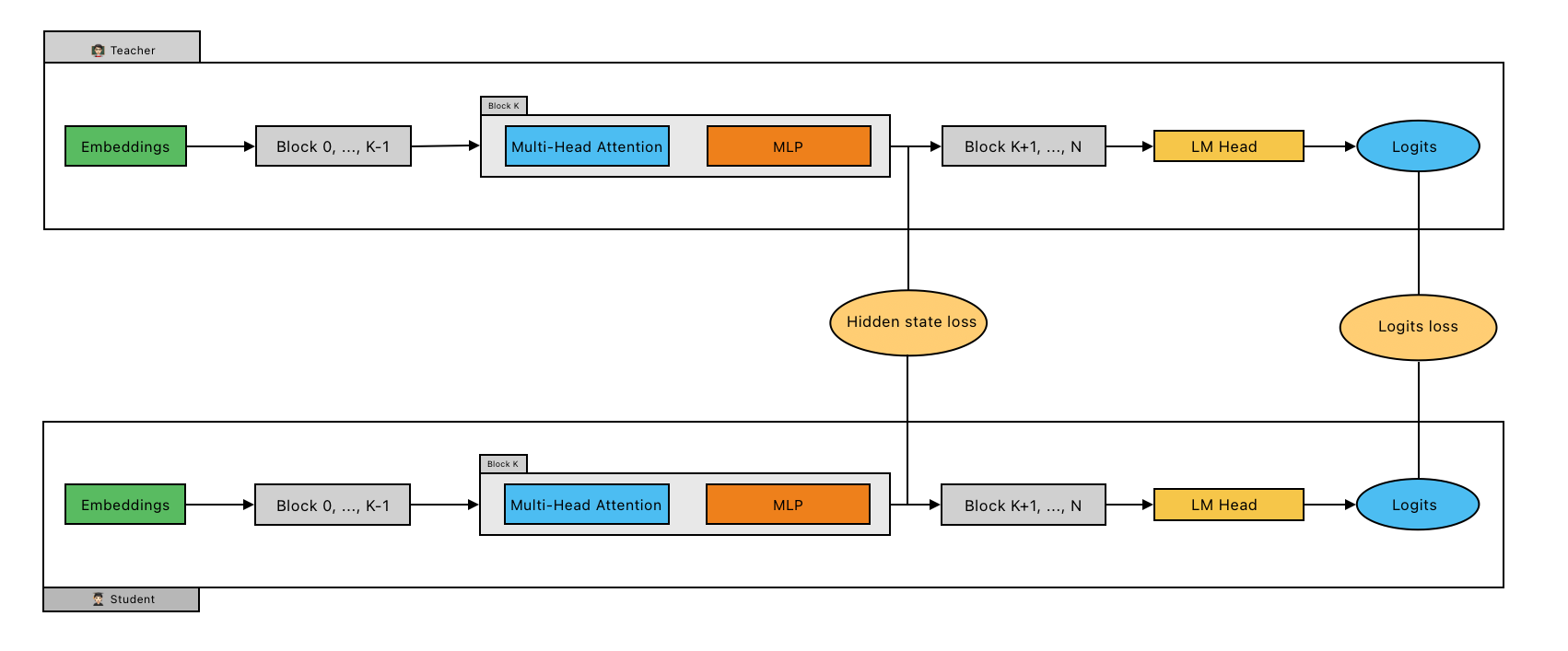}
    \caption{Fine-tuning with a combined loss function}
    \label{fig:logits+hs-ft}
\end{figure}

Each loss function targets a different aspect of knowledge transfer:
\begin{itemize}
    \item  KLDivLoss (logits/logprobs): emphasizes matching output probability distributions, important for tasks where semantic closeness of predictions matters.
    \item  MSELoss (hidden states): focuses on aligning internal representations, supporting transfer of deep structural and semantic features, especially valuable in multi-layer architectures.
    \item  Combined KLDivLoss and MSELoss (Fig. \ref{fig:logits+hs-ft}): simultaneously optimizes both the output probability distributions and internal representations. To balance their contributions, both normalized and unnormalized variants were evaluated.
\end{itemize}

\subsection{Experimental Results for Loss Functions}
All experiments used a version of \texttt{Qwen2.5-3B} with its two least important layers removed according to the pruning order identified as optimal from evaluations on representative datasets.
The results are shown in Table \ref{tab:ft-types}.
\begin{table}[!h]
    \centering
    \resizebox{\linewidth}{!}{
    \begin{tabular}{|l|c|}
    \hline
    \textbf{Model} & \textbf{Aggregate Score} \\
    \hline
    Original Qwen2.5-3B                          & 0.636 \\ \hline
    Pruned 2 Layers & 0.608 \\ \hline
    Pruned + HS MSE on All Layers  & 0.625 \\ \hline
    Pruned + HS MSE on Last Layers & 0.623 \\ \hline
    Pruned + HS MSE on Last Trainable Layer & 0.623 \\ \hline
    Pruned + HS MSE on Last Trainable + Last Layers & 0.624 \\ \hline
    Pruned + HS MSE on All Trainable Layers &  0.624 \\ \hline
    Pruned + KLDiv + MSE HS $\times$50 & 0.623 \\ \hline
    \textbf{Pruned + KLDiv Logits / 100 + MSE HS} & \textbf{0.626} \\ \hline 
    Pruned + KLDiv Log-Probs + MSE HS & 0.624 \\ \hline
    Pruned + KLDiv Log-Probs + MSE HS $\times$50  & 0.624 \\ \hline
    Pruned + KLDiv Log-Probs /100 + MSE HS & 0.624 \\ \hline
    Pruned + KLDiv Log-Probs / 100 + MSE HS $\times$10 & 0.623 \\ \hline
    Pruned + KLDiv / 100  + MSE HS  & 0.621 \\ \hline
    Pruned + KLDiv Reverse Teacher Log-Probs + MSE HS & 0.624 \\ \hline
    Pruned + KLDiv Teacher Log-Probs  & 0.623 \\ \hline
    Pruned + Reverse KLDiv Student Log-Probs & 0.613 \\ \hline
    Pruned + Reverse KLDiv Teacher Log-Probs & 0.624 \\ \hline
    Pruned + KLDiv & 0.608 \\ 
    \hline
    \end{tabular}
    }
    \vspace{5pt}
    \caption{Results of fine-tuning the pruned model with different loss functions. Best loss function is highlighted.
    MSE HS stands for MSELoss for Hidden States of last trainable and last model's layers}
    \label{tab:ft-types}
\end{table}

     As shown in Table~\ref{tab:ft-types}, the highest quality is achieved with a combined loss: \texttt{KLDivLoss} on logits (scaled by 1/100), along with the sum of \texttt{MSELoss} over the last trainable and last model layers.
     
     During the experiments, various combinations of hyperparameters were explored. The best balance between efficiency and quality was achieved with a learning rate of 1e-4 and a maximum sequence length of 2048. These hyperparameters were subsequently used to evaluate the proposed distillation approaches.
     


\subsection{Experiments with New Distillation Methods}
   $\qquad$Experiments were carried out on the \texttt{Qwen2.5-3B} model \cite{qwen2.5}, which has 36 layers and 3 billion parameters. The aim was to obtain a reduced model with 28 layers and 2.47 billion parameters, minimizing quality loss.

   For comparative analysis of new distillation methods with the original ShortGPT approach \cite{men2024shortgpt}, a reduced model was constructed using the ShortGPT method.

   \noindent
   Five different options for distilling the \texttt{Qwen2.5-3B} model are presented:

    \begin{enumerate}
        \item \textbf{ShortGPT}: Calculate the importance of all layers in the original model once using the Block Influence metric. Remove the least important layers based on this assessment, without recalculating importance after each removal.
        \item \textbf{Iterative Block Influence}: Calculate the importance of all layers using the Block Influence metric. Remove the least important layer, then recalculate importance after each removal to determine the next layer to prune.
        \item \textbf{Iterative Block Influence + Distillation}: Calculate the importance of all layers using Block Influence. After each removal, perform additional fine-tuning of the model, then recalculate importance considering changes introduced by fine-tuning.
        \item \textbf{Iterative Layer-wise Pruning}: Calculate the importance of all layers in the original model. After each removal, reassess importance by evaluating on the selected representative datasets, and proceed with the removal of the next least important layer.
        \item \textbf{Iterative Layer-wise Distillation}: Calculate the importance of all layers in the original model. After each removal, perform fine-tuning, then recalculate layer importance by evaluating on the selected representative datasets before continuing the pruning process.
\end{enumerate}

    The models obtained from these approaches are presented in Table \ref{tab:pruned8}.

    \begin{table}[h!]
    \centering
    \begin{tabular}{|l|c|}
    \hline
    \textbf{Model Variant}                        & \textbf{Aggregate Score} \\
    \hline
    Qwen2.5-3B (original)                         & 0.636 \\
    Qwen2.5-1.5B (original)                       & 0.601 \\
    Pruned - ShortGPT method                      & 0.175 \\
    Pruned – Iterative Block Influence            & 0.142 \\
    Pruned – Iterative Layer-wise Pruning         & 0.179 \\
    Pruned – Iterative Block Influence + FT       & 0.352 \\
    \textbf{Pruned – Iterative Layer-wise Distillation} & \textbf{0.574} \\
    \hline
    \end{tabular}
    \vspace{5pt}
    \caption{
        Comparison of 28-layer models obtained from \texttt{Qwen2.5-3B} and \texttt{Qwen2.5-1.5B}. Pruned models were derived from \texttt{Qwen2.5-3B} using various distillation strategies.
    }
    \label{tab:pruned8}
\end{table}

    Since Iterative Layer-wise Distillation approach demonstrated the best performance, it was further investigated and extended to a model with 24 layers and 2.161 billion parameters, so that this model could be classified as a ``1.5 billion'' class model.

    \begin{figure}[h!]
        \centering
        \includegraphics[width=1\linewidth]{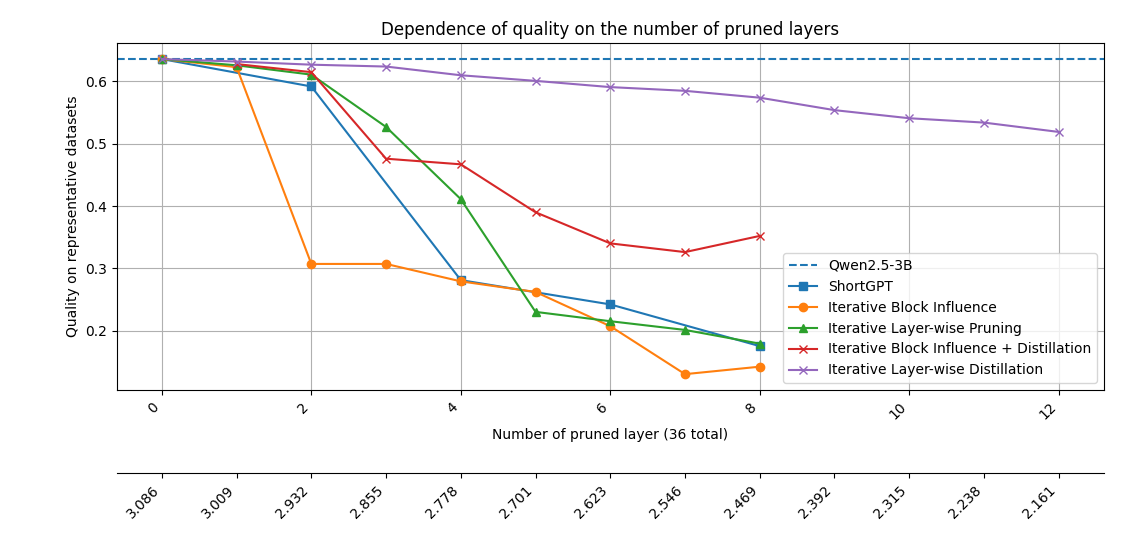}
        \caption{Results of various approaches on the \texttt{Qwen2.5-3B} model}
        \label{fig:distil-results}
    \end{figure}

    Figure \ref{fig:distil-results} illustrates how model quality depends on the number of layers removed. It is evident that models produced using the ShortGPT approach, as well as those employing pruning without a distillation step, experience a substantial quality drop - exceeding 70\% after the removal of eight layers. The Iterative Block Influence + Distillation method achieves approximately double the quality of these baseline approaches, yet the decline relative to the original model remains significant. In contrast, Iterative Layer-wise Distillation demonstrates the best performance among all methods considered. Notably, even after removing 12 layers using this approach, the resulting model achieves a score of \texttt{0.519} on representative datasets. Thus, eliminating one-third of the layers leads to only an 18\% reduction in quality compared to the original model.
    The model distilled using the Iterative Layer-wise Distillation approach is publicly available on Hugging Face\footnote{https://huggingface.co/kaengreg/Qwen2.5-2B-layerwise-distilled}.

    As a result of all experiments, statistics were collected (Fig. \ref{fig:rem-layers-stats}) on the layers removed in the approaches considered:
    \begin{figure}[h!]
        \centering
        \includegraphics[width=1\linewidth]{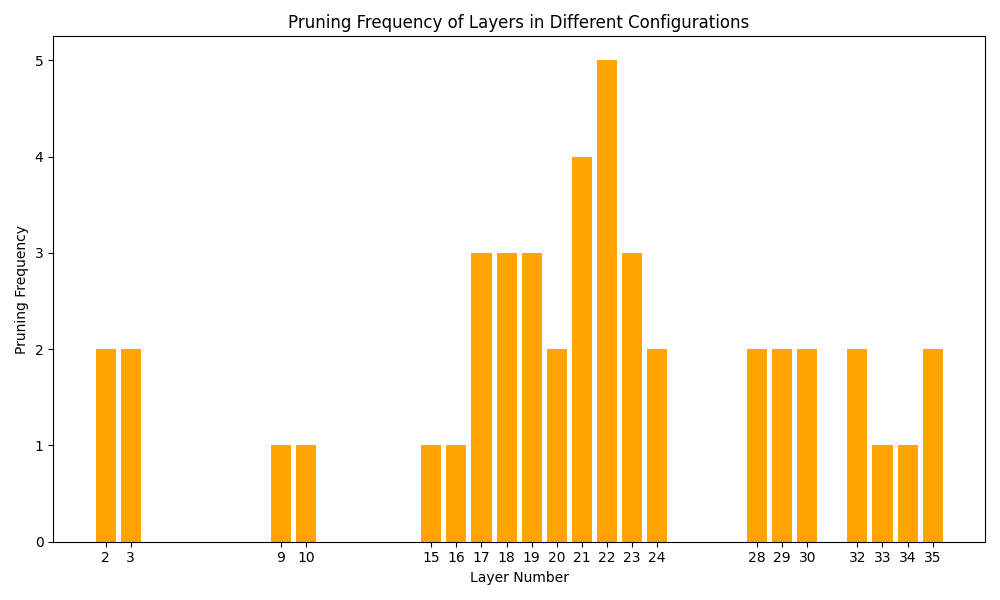}
        \caption{Statistics on removed layers for the considered approaches}
        \label{fig:rem-layers-stats}
    \end{figure}

   Analysis of the layer removal frequency (Fig. \ref{fig:rem-layers-stats}) indicates that layers 17–24 were most often identified as the least significant in our experiments. This is explained by the fact that they form intermediate representations, many of which partially duplicate each other. In transformer architectures, the middle part of the model performs transformations that contribute little to inference and can be removed without significant quality loss. Such layers serve as aggregation blocks, but do not directly participate in either feature extraction or final generation.

    In contrast, removing late (final) layers is highly undesirable. These layers play a key role in forming the final logits, i.e., they are responsible for converting internal representations into probability distributions over the vocabulary. Moreover, the uppermost layers concentrate the most semantically rich and contextually relevant features. Their removal severely degrades text generation and destroys the model's ability to produce meaningful responses.
    
\section{Conclusion}
This work explored and systematically evaluated distillation methods for large language models, with a focus on developing compact architectures that preserve high quality while reducing computational requirements. Central to the proposed method is an iterative, data-driven assessment of layer importance using representative datasets, combined with targeted fine-tuning on teacher logits and hidden states via KLDivLoss and MSELoss.

Experimental results on the Qwen2.5-3B model demonstrate that distillation strategies incorporating iterative, layer-wise fine-tuning after each pruning step consistently outperform approaches without fine-tuning. Notably, the Iterative Layer-wise Distillation method reduced the model from 36 to 28 layers (2.47B parameters) with less than a 10\% quality drop, and maintained stable performance even after further pruning to 24 layers. In contrast, methods lacking a distillation step resulted in severe degradation of model quality.

Analysis of the pruning process revealed that some intermediate layers (17–24) contribute little to inference, whereas the final layers are essential for preserving model performance. Selective fine-tuning of layers adjacent to those removed was found to be particularly effective for recovering lost dependencies. This approach holds significant promise for scaling down larger models, where such low-contribution layers are more prevalent, thereby enabling their efficient deployment in resource-limited environments.

In summary, combining iterative importance evaluation with targeted fine-tuning enables the construction of efficient, high-performing language models. Future research should further investigate the scalability and adaptability of these methods to a broader range of architectures and tasks.

\section*{Acknowledgements}
The research was supported by the Russian Science Foundation, project Nº 25-11-00191, https://rscf.ru/project/25-11-00191/.

The research was carried out using the MSU-270 supercomputer of Lomonosov Moscow State University.

\newpage
\section{Additional Results}

\begin{table}[h!]
\centering
\small
\begin{tabularx}{\linewidth}{|M{0.3\linewidth}|M{0.14\linewidth}|*{7}{X|}}
\hline
\textbf{Model Variant} & \textbf{Aggregate Score} &
\rotatebox{90}{\makebox[2.2cm][c]{enMMLU}} &
\rotatebox{90}{\makebox[2.2cm][c]{ruMMLU}} &
\rotatebox{90}{\makebox[3cm][c]{treeway\_abstractive}} &
\rotatebox{90}{\makebox[2.2cm][c]{cp\_doc\_ru}} &
\rotatebox{90}{\makebox[2.2cm][c]{cp\_para\_ru}} &
\rotatebox{90}{\makebox[2.2cm][c]{flores\_en\_ru}} &
\rotatebox{90}{\makebox[2.2cm][c]{flores\_ru\_en}} \\
\hline
\makecell[l]{Qwen2.5-3B\\(original)} & 0.636 & 0.68  & 0.565 & 0.241 & 1    & 0.99 & 0.445 & 0.534 \\
\makecell[l]{Qwen2.5-1.5B\\(original)}& 0.601 & 0.622 & 0.465 & 0.223 & 1    & 1    & 0.39  & 0.51  \\
\makecell[l]{Pruned - ShortGPT\\method} & 0.175 & 0.281 & 0.293 & 0.145 & 0.02 & 0.04 & 0.164 & 0.283 \\
\makecell[l]{Pruned - Iterative\\Block Influence} & 0.142 & 0.285 & 0.256 & 0.037 & 0.03 & 0.01 & 0.135 & 0.244 \\
\makecell[l]{Pruned - Iterative\\Layer-wise Pruning} & 0.179 & 0.594 & 0.467 & 0.004 & 0    & 0    & 0.01  & 0.178 \\
\makecell[l]{Pruned - Iterative Block\\Influence + FT} & 0.352 & 0.405 & 0.355 & 0.169 & 0.16 & 0.5  & 0.387 & 0.485 \\
\makecell[l]{\textbf{Pruned - Iterative}\\\textbf{Layer-wise Distillation}} & \textbf{0.574} & 0.562 & 0.458 & 0.204 & 0.95 & 0.98 & 0.381 & 0.481 \\
\hline
\end{tabularx}
\vspace{5pt}
\caption{Full comparison of 28-layer models obtained from \texttt{Qwen2.5-3B} and \texttt{Qwen2.5-1.5B}. Pruned models were derived from \texttt{Qwen2.5-3B} using various distillation strategies.}
\label{tab:full-results}
\end{table}

\bibliography{references}

\end{document}